\documentclass[runningheads,a4paper,10pt]{llncs}

\usepackage{amssymb}
\usepackage{graphicx}
\usepackage[margin=0.66in]{geometry}
\usepackage{hyperref}
\usepackage{palatino}
\usepackage{multicol}

\usepackage{url}
\urldef{\mailsa}\path{citlalli.gamez@gmail.com, alain.tremeau@univ-st-etienne.fr, ruven@c2rmf.cnrs.fr}
\newcommand{\keywords}[1]{\par\addvspace\baselineskip
\noindent\keywordname\enspace\ignorespaces#1}

\begin{document}

\mainmatter  

\title{Data Fusion of Objects using Techniques such as Laser Scanning, Structured Light and Photogrammetry for Cultural Heritage Applications}

\titlerunning{Lecture Notes in Computer Science - Computational Color Imaging}

%
%
\author{Citlalli G\'{a}mez Serna
\and Ruven Pillay \and Alain Tr\'{e}meau}
\authorrunning{Lecture Notes in Computer Science - Computational Color Imaging}

\institute{\begin{footnotesize}Hubert Curien Laboratory\\
Building F, 18 Rue Professeur Beno\^{\i}t Lauras \\
42000 St-Etienne, France \and
C2RMF\\
Palais du Louvre - Porte des Lions\\
14 Quai Francois Mitterrand\\
75001 Paris\\
France\end{footnotesize}}

%

\toctitle{Digitalization for Cultural Heritage Applications}
\tocauthor{}
\maketitle

\begin{abstract}
In this paper we present a semi-automatic 2D-3D local registration pipeline capable of coloring 3D models obtained from 3D scanners by using uncalibrated images. The proposed pipeline exploits the Structure from Motion (SfM) technique in order to reconstruct a sparse representation of the 3D object and obtain the camera parameters from image feature matches. We then coarsely register the reconstructed 3D model to the scanned one through the Scale Iterative Closest Point (SICP) algorithm. SICP provides the global scale, rotation and translation parameters, using minimal manual user intervention. In the final processing stage, a local registration refinement algorithm optimizes the color projection of the aligned photos on the 3D object removing the blurring/ghosting artefacts introduced due to small inaccuracies during the registration. The proposed pipeline is capable of handling real world cases with a range of characteristics from objects with low level geometric features to complex ones. 

\end{abstract}

\begin{multicols}{2}

\section{Introduction}

Digitization of cultural heritage objects has gained great attention around the world due to the importance and awareness of what they represent for each culture. Researchers have been trying to achieve the same goal: capturing a 3D digital representation together with its color information to be able to pass them down safely to future generations.

The recovery and generation of the 3D digital representation requires high geometric accuracy, availability of all details and photo realism \cite{el2002detailed}. Any single 3D imaging technique is unable to fulfil all of these requirements and the only way to solve this problem is through the fusion of multiple techniques.

There have been a number of recent studies which have tried to map automatically, semi-automatically or manually a photorealistic appearance onto a 3D model. Some of these have used only photogrammetry \cite{pollefeys2000automated},\cite{fitzgibbon1998automatic}, which provides poor geometric precision. However for cultural heritage applications, especially for conservation, a high density of the 3D point cloud is needed. In order to satisfy the demanding needs of cultural heritage, the combination of both photogrammetry and range scans \cite{bernardini2001high,liu2006multiview,dellepiane2008mapping} have been considered. These approaches generally start by computing an image-to-geometry registration, followed by an integration strategy. The first one generally seeks to find the calibration parameters of the set of images, while the second tries to select the best color for each of the images.

There has been research focusing on improving the alignment in all the images \cite{liu2005automatic,dellepiane2013,li2009automatic} (global registration). However, the visual results show significant blurring and ghosting artefacts. Others have proved that a perfect global registration is not possible because the two geometries come from different devices and consequently the only solution available is to consider a local registration refinement \cite{dellepiane2012flow,eisemann2008floating,takai2010harmonised}. 

This paper proposes a solution for a full end-to-end pipeline in order to process data from different acquisition techniques to generate both a realistic and accurate visual representation of the object. Our solution recovers the 3D dimension from 2D images to align the 3D recovered object with a second more geometrically accurate scan. The input 2D images are enhanced to improve the feature detection by the Structure from Motion algorithm (SfM) which provides the position and orientation of each image together with a sparse 3D point cloud. The idea behind the 3D reconstruction is to perform the alignment in 3 dimensions through the Scale Iterative Closes Point (SICP) algorithm obtaining the transformation parameters to be applied in the extrinsic ones of the cameras. Even though, the alignment is performed minimizing the distance between both 3D models, it is approximate for different reasons (sparseness, noise) and a local registration refinement is needed. In the last stage of our pipeline, color projection, an algorithm to correct the local color error displacement is performed. Our local correction algorithm works in an image space finding the correct matches for each point in the 3D model is deviated from image to image.



\section{Related Work}
The main related issues taken into account in our pipeline can be divided into 3 major fields: (1) 2D/3D registration, (2) color projection, and (3) registration refinement process. The important related work in these fields is outlined below.

\subsection{2D/3D Registration}
Image to Geometry registration consists of registering the images with the 3D model defining all the parameters of the virtual camera (intrinsic and extrinsic) whose position and orientation gives an optimal inverse projection of the image onto the 3D model. 

Numerous techniques exist and a number of different ways exist to try to solve this problem. The methods can be classified into (i) manual, (ii) automatic or semi-automatic depending mainly on matches or features. In the (i) manual methods the registration is performed manually selecting correspondences between each image and the 3D geometry. This technique is often used for medical applications \cite{liu19983d}. Others instead, have used features in order to automate the process, but finding consistent correspondences is a very complex problem. Due to the different appearance of photographs and geometric models, (ii) automatic methods are limited to some specific models and information. For example, line features are mostly used for urban environments \cite{liu2005automatic},\cite{musialski2013survey}; and silhouette information is used when the contour of the objects is visible in the images and the 3D model projected onto an image plane \cite{ip1996constructing,lensch2000automated,lensch2001silhouette}. 

Nevertheless there are 3D scanners which provide also reflectance images and the registration is performed in a 2D space \cite{ikeuchi2007great},\cite{kurazume2002simultaneous}. On the other hand, some authors perform their registration in a 3D space reconstructing the 3D object from the 2D images and aligning both 3D objects \cite{liu2006multiview},\cite{li2009automatic},\cite{corsini2013fully}. This procedure is carried out in two steps: (1) 3D reconstruction and (2) point cloud alignment. Through the widely used Structure from Motion technique (SfM), a 3D reconstruction and intrinsic and extrinsic camera parameters are recovered without making any assumptions about the images in the scene. The registration is usually performed by selecting correspondences \cite{li2009automatic} that minimize the distances between a set of points.

Our work is based on SfM approach and the use of the SICP algorithm \cite{zhu2010robust} to register both point clouds with the only constraint being to locate them relatively close to each other.

\subsection{Color Projection}
Once the images are registered onto the 3D model, the next step is to exploit the photo-realistic information (color, texture) obtained by an optical sensor, together with the geometric details (dense point cloud) obtained by some type of 3D scanner (laser scanner, structured light). The aim is to construct a virtual realistic representation of the object. 

As a point in the 3D model projects onto different images and images may possess some artifacts (highlights, shadows, aberrations) or small calibration errors, the selection of the correct color for each point is a critical problem. In order to deal with this task, research has been based on different solutions, each one with its own pros and cons.

\subsubsection{Orthogonal View}
In \cite{lensch2000automated},\cite{callieri2002reconstructing}, the authors assign the best image to each portion of the geometry. This assignment relies on the angle between the viewing ray and the surface normal. As the color of a group of 3D points comes from one image, seams are produced when adjacent groups are mapped with different images and also artifacts such as differences in brightness and specularities are visible. Even though some research has dealt with the seams by smoothing the transitions \cite{lensch2000automated}, important and critical detail can be lost.\\

\subsubsection{Weighting Scheme}
In these kind of approaches \cite{bernardini2001high},\cite{li2009automatic},\cite{callieri2008masked}, an specific weight is assigned to each image or to each pixel in the images according to different quality metrics. The metrics vary between authors considering visible points, borders or silhouettes, depth \cite{li2009automatic},\cite{callieri2008masked} or the distance to
the edge of the scan \cite{bernardini2001high}.  All these methods, in comparison with orthogonal view, are able to eliminate the artifacts previously mentioned but instead introduce blurring/ghosting when the calibration of the images is not sufficiently accurate.

\subsubsection{Illumination Estimation}
Alternatively, approaches such as \cite{dellepiane2010improved} attempt to make an estimation of the lighting environment. This approach is able to remove possible illumination artifacts presented in the images (shadows/highlights). Unfortunately, in real scenarios it is difficult to accurately recover the position and contribution of all the light sources in the scene.

Due to the evaluation criteria used and advantages provided by all of these approaches, a weighting procedure was selected as the best option for our work. We used the approach by Callieri et al. \cite{callieri2008masked}  because of its robustness, availability and the good results obtained with it from our data set.

\subsection{Registration Refinement}
Since the data comes from 2 different devices and the geometry and camera calibration is imprecise after the 2D/3D registration; blurring or ghosting artifacts appear once the color projection is performed. In order to remove them, a global or local refinement is necessary.   

\subsubsection{Global Refinement}
Some approaches try to correct the small inaccuracies in a global manner \cite{liu2005automatic,dellepiane2013,li2009automatic},\cite{lensch2000automated} by computing a new registration of the camera parameters according to the dense 3D model obtained with the scanner. The goal is to distribute the alignment error among all the images to minimize the inaccuracies and improve the quality of the final color of the model. Unfortunately as the registration is mostly based on features, an exact alignment will not be possible due to image distortions or low geometric features. Nevertheless even if the global alignment refinement finds the best approximate solution, the matches will not be exactly the same. As a consequence blurry details (ghosting effects) will appear after the color projection \cite{dellepiane2012flow}, especially when the color details are in areas with low geometric features. The only straightforward solution to correct these small inaccuracies, is to perform a local refinement.

\subsubsection{Local Refinement}
A number of studies have been carried out based on local refinements which locally analyze the characteristics of each point and try to find the best correspondence in the image series \cite{dellepiane2012flow,eisemann2008floating,takai2010harmonised}.  Finding  these correspondences locally has been addressed in the literature by using optical flow. Some studies have computed dense optical flow \cite{dellepiane2012flow},\cite{eisemann2008floating} but the results depend on the image resolution and the amount of mismatch (displacement) together with the computational power available. On the other hand, others instead of working in the image space, have tried to optimize the 3D geometry in order to deform textures more effectively \cite{takai2010harmonised}. As our 3D geometry cannot be modified these kind of approaches are not feasible for our purpose. 

Computing dense optical flow in our datasets was impossible due to relatively high resolution of the images, e.g. 4008$\times$5344 pixels compared to the  1024$\times$768 pixels used in the literature in \cite{eisemann2008floating}. For this reason we decided to use sparse optical flow to compute the color for each point in the 3D geometry limiting the number of images to the best three, evaluated according to the quality metrics of Callieri et al. \cite{callieri2008masked}.


\section{Data Fusion}
\label{sec:DataFusion}
\begin{figure*}[h]
     \centering
     \includegraphics[width=0.9\linewidth]{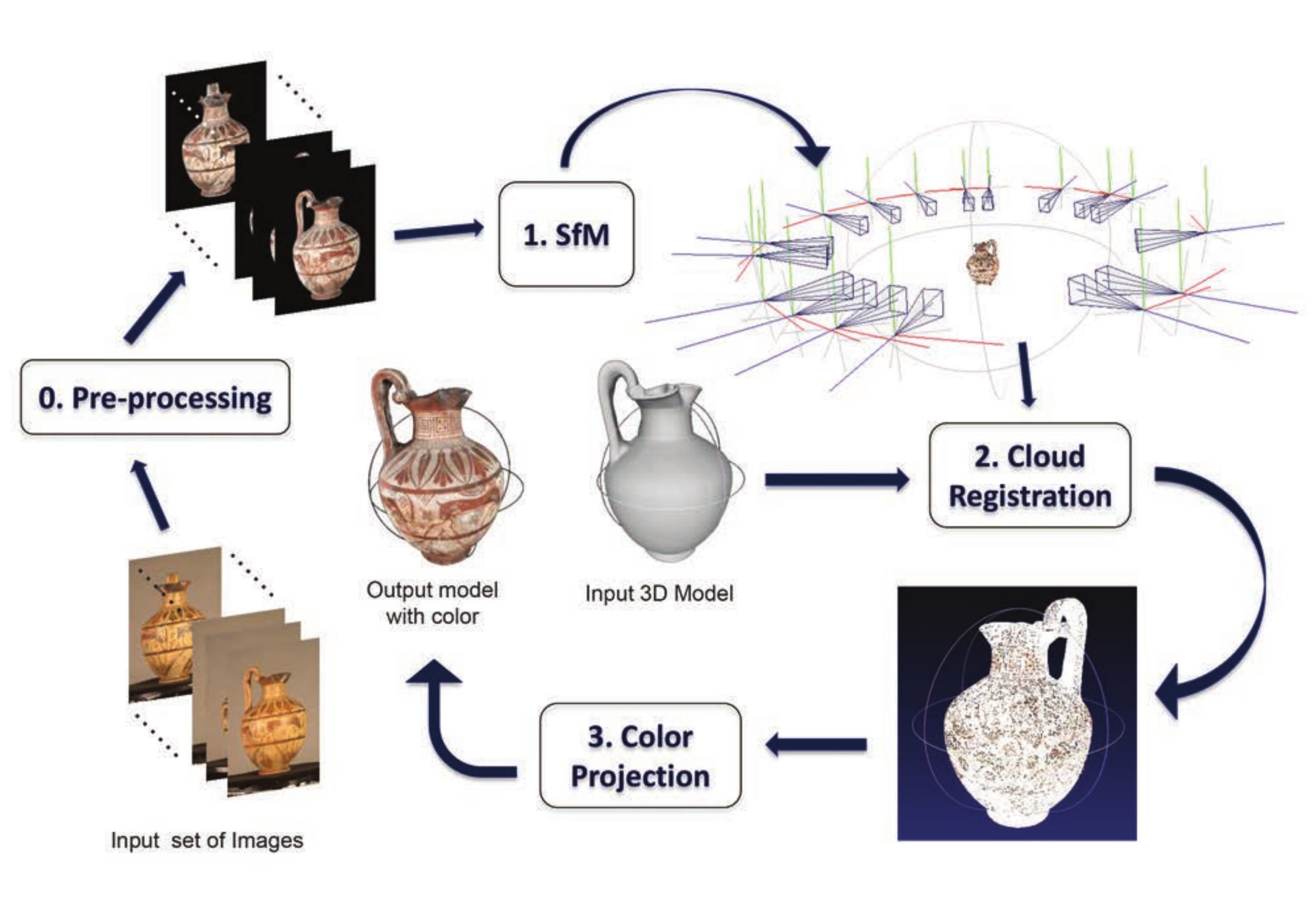}
     \caption[Pipeline overview]{General overview of the pipeline.}
     \label{fig:Pipeline}
 \end{figure*}

Our goal is to fuse the information provided by the two different devices (3D scanner and 2D camera) in order to recreate a high resolution realistic digital visualization with both very accurate geometric and visual detail. The procedure to achieve this result needs to take into account various problems which will be solved in essentially four main stages (see figure \ref{fig:Pipeline}): (1) Image preprocessing, (2) Camera calibration through Structure from Motion, (3) Cloud registration to align the images to the geometry, and (4) Color projection which involves the most correct images to project the color onto the 3D geometry. The whole process is designed to consider as input a set of uncalibrated images and a dense 3D point cloud or a 3D triangulated mesh. By uncalibrated images we refer to images in which the intrinsic and extrinsic camera parameters are unknown.

\end{multicols}

\begin{multicols}{2}

\subsection{Stage 0: Image Preprocessing}
Even though a number of studies have used a set of uncalibrated images to perform camera calibration and 3D reconstruction through some Structure from Motion algorithm \cite{liu2006multiview},\cite{li2009automatic},\cite{corsini2013fully},\cite{gallo20143d}, very few have considered a preprocessing step \cite{gallo20143d}. 

This stage is performed in order to improve the camera calibration procedure (Stage 1) and consequently obtain more accurate camera parameters together with a better 3D representation.

Three preprocessing steps were considered. The first two had already been applied by the C2RMF to their data sets and the introduction of a third preprocessing step also enabled an improvement for the next stage.

\begin{enumerate}
\item Color calibration. Performed to accurately record the colorimetric appearance of the object in the set of color images and to eliminate mis-matches caused by varying lighting conditions. In order to calibrate, a color chart is used during the image acquisition to determine a color transformation between the captured values and the reference color target.

\item Background subtraction. As a SfM procedure is based on feature matching, features will be detected in the foreground as well as in the background. In order to avoid the reconstruction of unwanted points (outliers) and have a clean 3D object, the background was removed manually. There are many segmentation techniques available in the literature \cite{pal1993review} but in order to be precise the manual method was considered by the C2RMF.

\item Image enhancement.  Through  histogram equalization, we enhance the image contrast in order to find a larger number of features and generate more 3D points in the next stage. The original principle applies to gray-scale
images, but we used it in color, changing from RGB to the HSV color space and equalizing the Value (V) channel in order to avoid hue and saturation changes
\cite{naik2003hue}. This step is very useful especially when the object lacks  texture details. The same idea was exploited in \cite{gallo20143d} with a Wallis filter.
\end{enumerate}

\subsection{Stage 1. Camera Calibration and 3D Reconstruction}
The second stage of our pipeline consists of a self-calibration procedure. It is assumed that the same camera, which is unknown, is used throughout the sequence and that the intrinsic camera parameters are constant. The task consists of (i) detecting feature points in each image, (ii) matching feature points between image pairs, and (iii) running an iterative robust SfM algorithm to recover the camera parameters and a 3D structure of the object.

For each image, SIFT keypoints are detected \cite{lowe2004distinctive} to find the corresponding matches using approximate nearest neighbors (ANN) kd-tree package from Arya et al. \cite{arya1998optimal} and the RANSAC algorithm \cite{fischler1981random} to remove outliers. Then a Structure from Motion (SfM) algorithm \cite{snavely2006photo},\cite{snavely2008modeling} is used to reconstruct a sparse 3D geometry of the object and obtain the intrinsic (i.e. focal length, principal point and distortion coefficients) and extrinsic (i.e. rotation and translation) camera parameters.

In order to achieve  a more geometrically complete surface of the 3D object,  Clustering Views from Multi-view Stereo (CMVS) \cite{furukawa2010towards} and Patch-based Multi-view Stereo (PMVS) \cite{furukawa2010accurate} tools are used. This aims to increase the density of the 3D geometry and be able to obtain a more precise parameter estimation during the cloud registration (stage 2).

\subsection{Stage 2. Cloud Registration}
After the 3D geometry obtained with the SfM algorithm and from the 3D scanner, a 3D-3D registration process is performed. As both points clouds possess different scales and reference frames, we will need to find the affine transformation that determines the scale (s), rotation (r) and translation (t) parameters which aligns better both 3D geometries. 

Usually a 3D-3D registration refers to the alignment between multiple point clouds scanned with the same device. Algorithms like Iterative Closest Point (ICP) \cite{zhang1992iterative} and 4 Point Congruent Set \cite{aiger20084} evaluate the similarity and minimize the distance between the 3D point clouds considering only the rotation and translation parameters. But when a scale factor is involved it can be solve separately  or together from the registration procedure. 

Calculating a bounding box for both 3D geometries and applying the ratio found between them seems to solve the scale problem, but if some outliers are present in one of the geometries the result will not be correct. Therefore Zhu et al. \cite{zhu2010robust}, extended the Iterative Closest Point algorithm to consider also the scale transformation (SICP), introducing a bidirectional distance measurement into the least squared problem. This algorithm works as follows: (i) define a target (fixed) and source (transforming) point clouds, which will be the scanned and reconstructed point clouds respectively in order to bring the camera parameters from the image space to the real object space; and (ii) perform iteratively the distance error minimization using the root mean square error (RMSE), until the best solution is found. The output is a set of 3D points aligned to the object coordinate system (real scale) by means of a 3$\times$3 rotational matrix, 3$\times$1 scale matrix and vector indicating the translation in X, Y and Z axis.

\subsection{Stage 3. Color Projection}
Color projection is the last and the core of our proposed pipeline. The aim is to project accurate color information onto the dense 3D model to create a continuous visual representation from the photographic image set. 

Selecting the best color is not an easy task; first because a single point in the 3D geometry is visible in multiple images and those may present  differences in illumination. Secondly small errors in the camera parameters cause small misalignments between the images, and in consequence, a point which projects onto a specific are in one image plane will project onto a slightly different area in another one. This can  result in different colors for each 3D point projected from several 2D images.

In order to address this problem some research based on the color selection from the most orthogonal image for a certain part of the 3D geometry \cite{lensch2000automated},\cite{callieri2002reconstructing} generating artifacts like highlights, shadows and visible seams.

Others project all images onto the 3D mesh and assign some weight to each image, as, for example, in \cite{bernardini2001high},\cite{li2009automatic},\cite{callieri2008masked}, which can remove artifacts that the orthogonal view is not capable of removing, but this can produce some ghosting artifacts when the alignment is not perfect.

In order to deal with less artifacts, we consider the approach based on Callieri et al. \cite{callieri2008masked}, which weights all the pixels in the images according to geometric, topological and colorimetric criteria.

The general procedure to perform the color projection in our pipeline takes into account two steps: (1) a color selection considering weights assigned according to the quality of each pixel, and (2) a local error correction in the image space in order to produce sharp results.

\subsubsection{Weighted Blending Function}
\begin{figure*}[h]  
      \centering
      \includegraphics[width=0.8\textwidth]{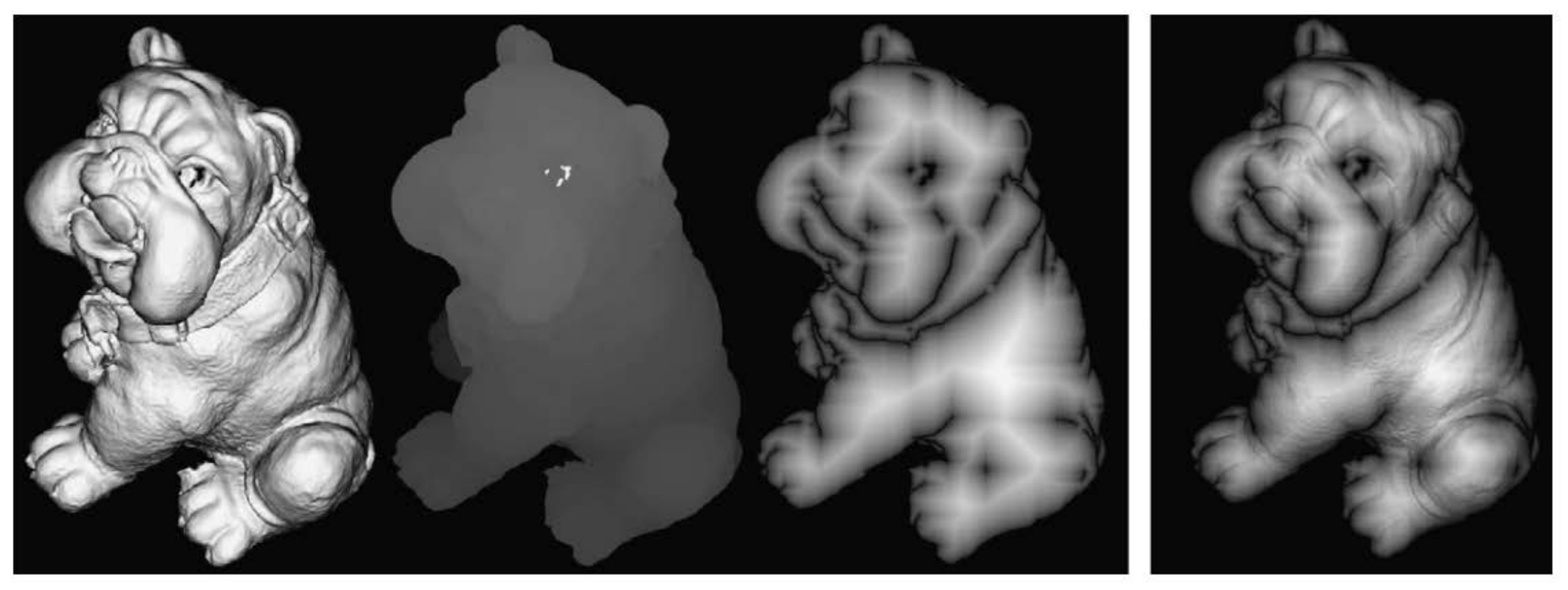}
      \caption[Example of weighting masks]{Example of weighting masks \cite{callieri2008masked}. From left to right: Angle mask, Depth mask, Border mask. Right-most, the combination of the previous three masks. For illustration purposes the contrast for the depth and border mask have been increased.}
      \label{fig:qualityMasks}
\end{figure*}

Through the approach by Callieri et al. \cite{callieri2008masked} it is possible to compute the weights for each 3D point in the number of images they are visible. The three metrics are based on: angle, depth and borders. For each of these metrics, a mask is created which has the same resolution as the original image. The aim is, therefore, to create a unique mask combining the three masks through multiplication. The result is a weighting mask for each image that represents a per-pixel quality (see an example in figure \ref{fig:qualityMasks}).

\end{multicols}

\begin{multicols}{2}

Once the weights are defined for each image, and knowing the camera parameters obtained in Stage 1, it is possible to perform a perspective projection from the 3D world onto an image plane. This projection allows us to know the color information for each 3D point in the geometry. The final color for each point is a weighted mean obtained by multiplying the RGB values from the pixels with their respective weights.

\subsubsection{Local Error Correction}
The results obtained projecting the color information with the quality metrics of Callieri et al. \cite{callieri2008masked} into the 3D geometry, generated blurring/ghosting effects in some parts of the mesh. These problems appear due to small inaccuracies introduced in the image-to-geometry registration \cite{dellepiane2012flow}.

\begin{figure*}[h]
      \centering
      \includegraphics[width=0.8\linewidth]{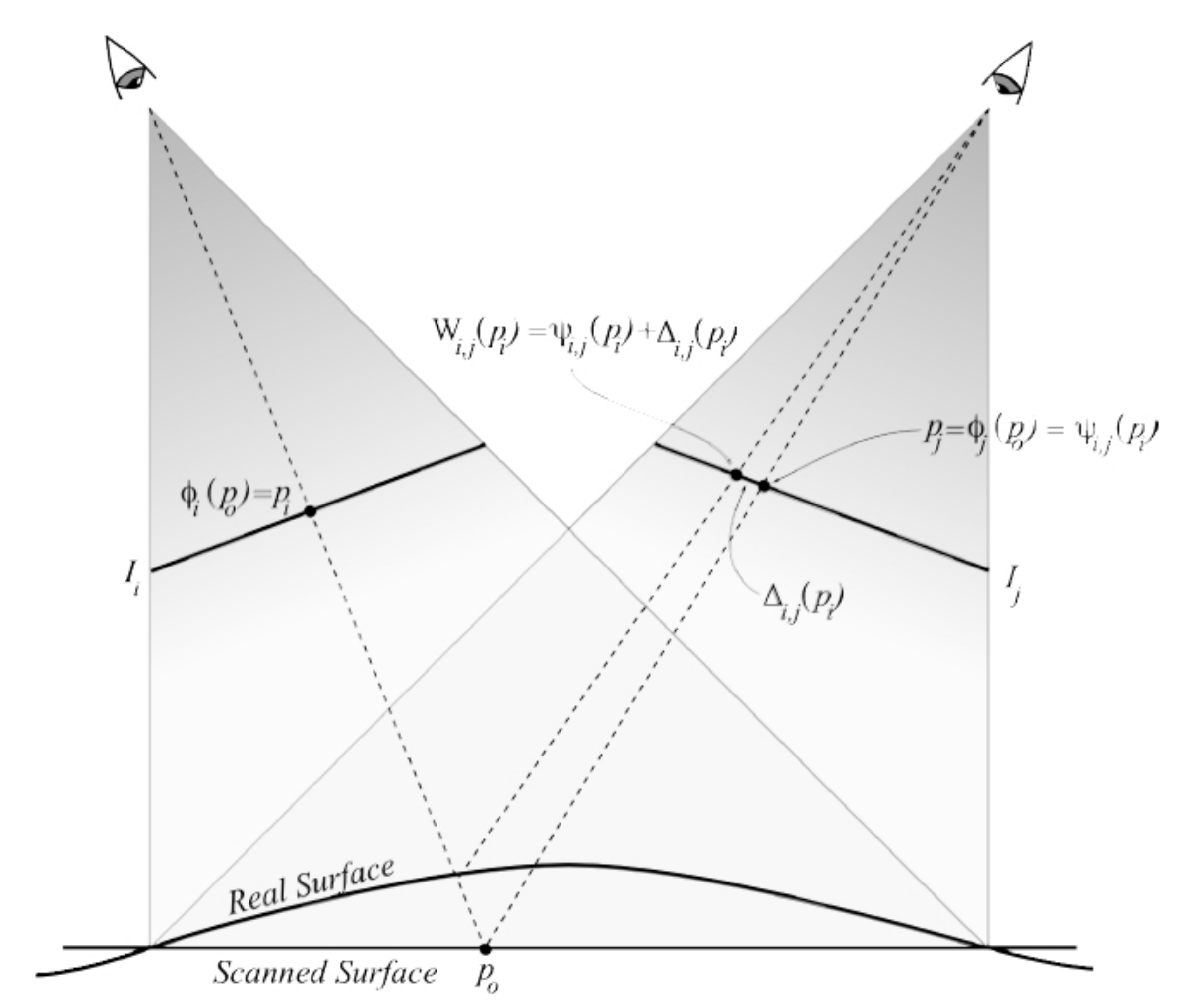}
      \caption[Graphic representation of the local displacement.]{Graphic representation of the local displacement defined by Dellepiane et al. \cite{dellepiane2012flow} where $p_o$ is the original point located in the scanned surface geometry; $\phi_i(p_o)$ represents the projection from the 3D world int a 2D image plane; $\psi_{i,j}(p_i)$ is the relation between corresponding points on different images; $\Delta_{i,j}(p_i)$ is the necessary displacement required to find the correct matches; and $W_{i,j}(p_i)$ is the warping function necessary to find the correspondent point in the second image plane.}
      \label{fig:displacement}
\end{figure*}

Research such as that by  \cite{dellepiane2012flow,eisemann2008floating,takai2010harmonised},\cite{pighin2006synthesizing} have considered these kind of artifacts; but their origin, is explained by Dellepiane et al. \cite{dellepiane2012flow} in figure \ref{fig:displacement}.

\end{multicols}

\begin{multicols}{2}

\begin{figure*} [h] 
      \centering
      \includegraphics[width=0.9\linewidth]{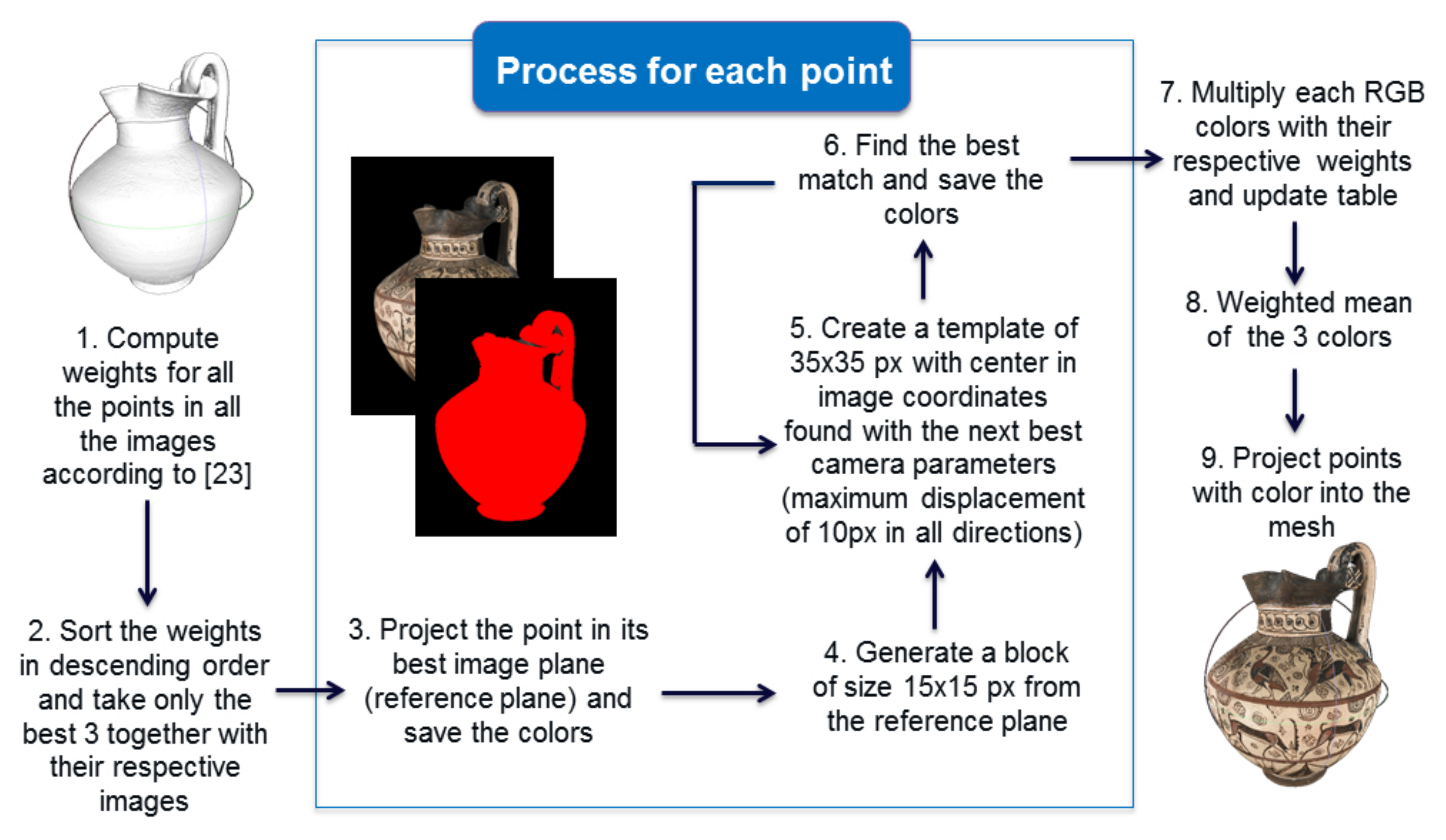}
      \caption[Illustration of the local error estimation procedure.]{Illustration of the local error estimation procedure.}
      \label{fig:localErrorAlgo}
\end{figure*}

The simplest way to correct these inaccuracies which generate the blurring artifacts, consists of finding for each 3D point, the local displacement in the best 3 image planes where it is visible. This local error estimation algorithm, based on \cite{dellepiane2012flow}, is performed through a template matching algorithm shown in figure \ref{fig:localErrorAlgo}. 

The reason for considering only the best three images for each point, instead of all where it is visible, is to speed up the process in the final color calculation. Instead of computing {\it (n-1)p} evaluations, we reduce them to {\it (3-1)p} where n is the number of images and p the points. Dellepiane et al. \cite{dellepiane2012flow} affirmed that three images are enough to correct illumination artifacts.

\end{multicols}

\begin{multicols}{2}

The size of the block and template was defined according to some experimental evaluation performed on the highest resolution dataset (FZ36147). Normally if the cloud registration step in Stage 2 is accurate enough, the different 3D points projected in the image plane will not be so far from each other. For this reason the same parameters can be applied to lower resolution images, but they cannot be considered for even higher ones. The most straightforward solution is to tune the parameters depending on the image resolution and the 3D cloud registration output.

The matching procedure is done on a pixel-by-pixel basis in a Luminance-Chrominance color space. The conversion of the RGB values into YCbCr color space was performed directly with the built-in Matlab function  'rbg2ycbcr' and the similarity measurement mean square error (MSE) was defined considering also changes in brightness for each block by subtracting the average value in each channel. Through this subtraction we account for big changes in illumination between images. The notation is the following:

\begin{equation}
MSE= \frac{1}{N^2} \sum_{i=0}^{N-1} \sum_{j=0}^{N-1} ( (S_{ij}-\overline{S}) - (T_{i,j}-\overline{T}) )^2
\end{equation}   

where N is the total number of pixels in each block, S is the block in the source/reference image, T is the block inside the template of the target image, $\overline{S}$ and $\overline{T}$ are the mean values of their respective channels. At the end the error with the minimum value is considered as the best match.

\begin{equation}
Error=\frac{MSE_Y+MSE_{Cb}+MSE_{Cr}}{3}
\end{equation}

In the case where there is more than one block matching the same criterion, a comparison of the colors from the center points will decide which block in the template is the closest to the block from the reference image. 

When the three RGB color values are found for each point, we proceed with the multiplication of them with their respective weights to average the results and assign final color values to each point in the 3D geometry.


\section{Experimental Results}
In this section we present experiments performed on real data from the C2RMF with scans from objects from the Department of Roman and Greek Antiquities at the Louvre museum in order to assess the performance of the proposed pipeline. The data had been captured at different times using different equipment. Each object had data from a structured light scanner and a set of color images used for photogrammetry.

The 2 data sets (FZ36147 and FZ36152) contain information with different qualities and sizes and a small description of the datasets used, is listed below together with a brief explanation of the criteria used for a visual quality evaluation.

\begin{itemize}
{\bf \item Dataset FZ36147. } This Greek vase, an Oenochoé from around 610BC, contains 1,926,625 points (pts) and 35 high resolution images (4008$\times$5344 pixels).The images were acquired under an uncalibrated setup, but our method was capable to remove the lighting artifacts and preserve details in its decorations. For the final evaluation of our proposed local error estimation algorithm implemented as part of the color projection procedure (stage 3), three small patches selected manually from the 3D geometry were extracted. Each patch was carefully selected according to visual details where mis-registration of the camera parameters led to blurring artifacts.  

{\bf \item Dataset FZ36152. } This Greek vase, an Oenochoé from between 600-625BC, represented by a 3D model which contains 1,637,375 points and 17 images of resolution 2152$\times$3232 pixels. With this dataset, the registration in the second stage of our pipeline, is not accurate enough to avoid blurring effects which appear in the whole 3D geometry. The local error correction in our method, brings sharp results in the three patches extracted in the same way as in the previous dataset.

\end{itemize}

Due to the fact that the images of the datasets are uncalibrated (no ground truth is available) only qualitative, meaning visually, evaluations were performed, as found also in the state of the art \cite{dellepiane2012flow}, \cite{corsini2013fully}. 

The reason for performing the evaluation only in small patches, refers to the high density of points each 3D geometry contains and the programming language used for the implementation (CPU programming).

\begin{figure*}
      \centering
      \includegraphics[width=0.8\linewidth]{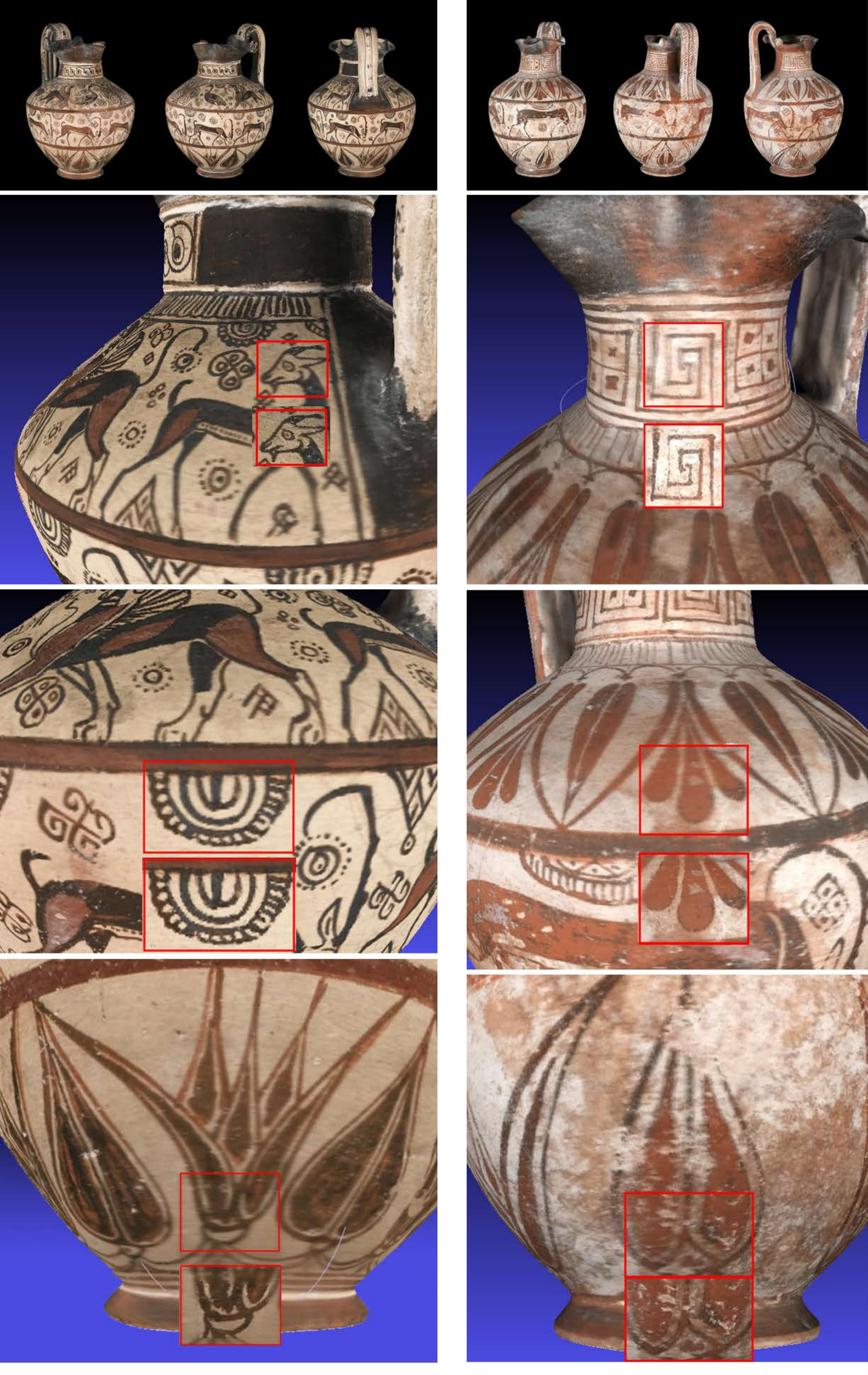}
      \caption[Final color projection.]{Final color projection in datasets from left to right FZ36147 and FZ36152. In the first row some of the original images are illustrated; second to fourth represents the 3 patches used for the direct color projection with Callieri et al. approach \cite{callieri2008masked} and the local error correction results for each of them (down red squares).}
      \label{fig:LocalResults}
\end{figure*}

Our algorithm, implemented in stage 3, corrects the small errors in the projected image planes, converging to good results regardless the initial rough alignment obtained during stage 2. Figure \ref{fig:LocalResults} shows the results of the color projection once the small errors are corrected. The quality of the appearance in the new projections (down red squares) is improved, removing the unpleasant blurring artifacts. Table \ref{tab:FinalResults} shows a summary of the characteristics of the datasets used, together with the patches evaluated and their corresponding computational time.

\end{multicols}

\begin{table*}
  \centering
  \begin{tabular}{|c|c|c|c|c|c|}
    \hline
    {\bf Dataset}	&	{\bf 3D model size}	&	{\bf N. of images}	&	{\bf Patch}	&	{\bf S. Patch}	&	{\bf Computational}	\\
    	&		&	{\bf (Resolution)}	&		&		&	{\bf Time}	\\
    \hline
    FZ36147	&	1,926,625 pts	&	35 (4008$\times$5344)	&	Up	&	4049 pts	&	2 hrs 30 min	\\
    	&		&		&	Middle	&	4834 pts	&	3 hrs 3 min	\\
    	&		&		&	Down	&	3956 pts	&	6 hrs 40 min	\\
   \hline
    FZ36152	&	1,637,375 pts	&	17 (2152$\times$3232)	&	Up	&	4750 pts	&	3 hrs 10 min	\\
    	&		&		&	Middle	&	4903 pts	&	2 hrs 46 min	\\
    	&		&		&	Down	&	6208 pts	&	3 hrs 8 min	\\
 
    \hline
  \end{tabular}
   \caption[Overview of tests performed with our Local error estimation algorithm]{Overview of tests performed with our Local error estimation algorithm.}
  \label{tab:FinalResults}
\end{table*}

\begin{multicols}{2}

The time required to perform the local error estimations, depends on the amount of 3D points the geometry has, and on the displacement found for every projected point in the 2D image plane. If the density increases, the computational time will be higher.

\subsubsection{Discussion}

\end{multicols}

\begin{figure*} [h]
         \centering
         \includegraphics[width=1\linewidth]{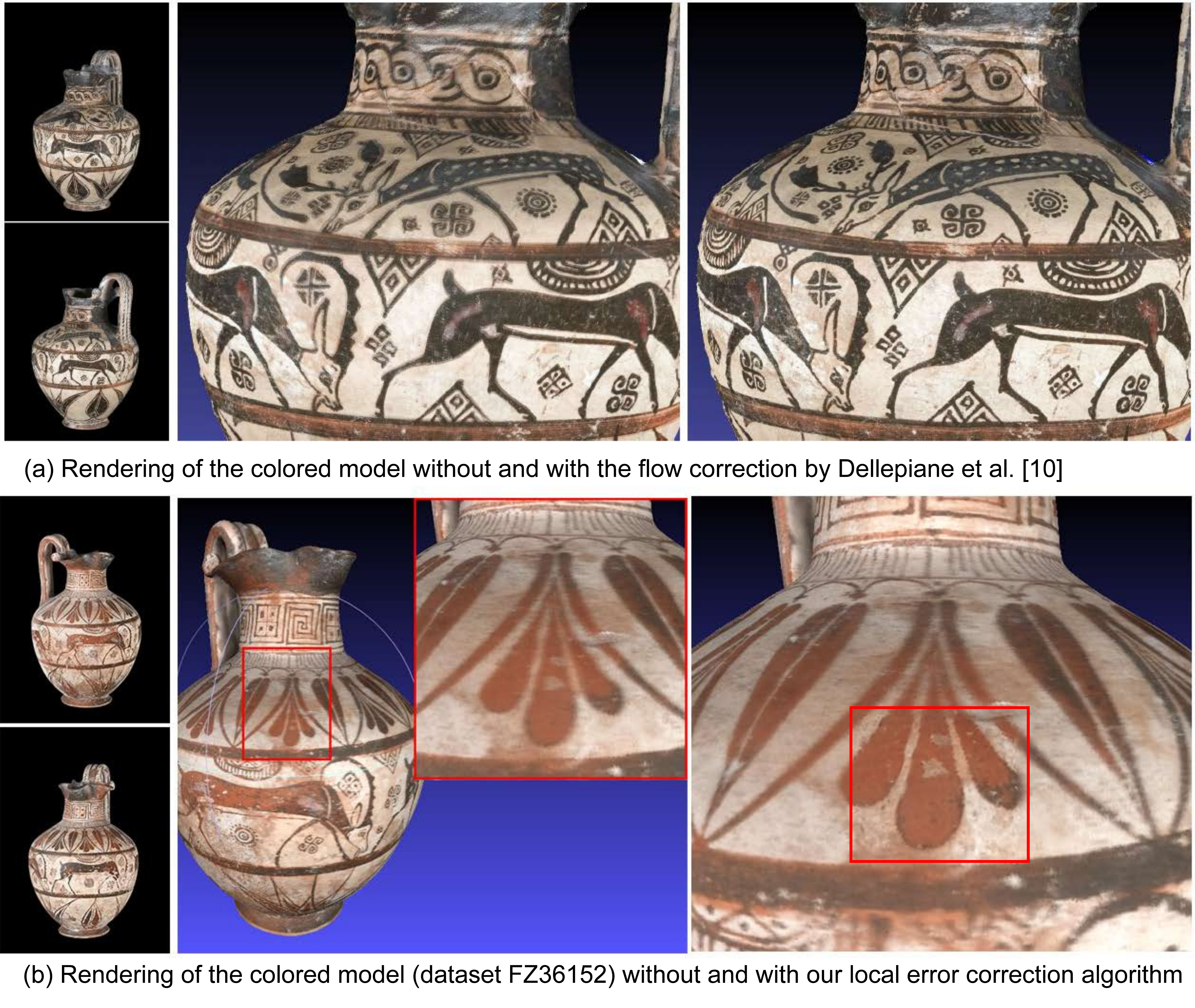}
         \caption[Final color projection comparison with the State of the Art.]{Final color projection comparison with the State of the Art.}
         \label{fig:comparisonStateOfArt}
\end{figure*}

\begin{multicols}{2}

A visible comparison with the state of the art \cite{dellepiane2012flow} is presented in figure \ref{fig:comparisonStateOfArt}. Dellepiane et al. also evaluated their method with one dataset from the Louvre museum with different resolution characteristics. The implementation in \cite{dellepiane2012flow} is based on dense optical flow and GPU programming for which really high resolution images are a disadvantage. The maximum resolution tested by Dellepiane et al. \cite{dellepiane2012flow} was 3000$\times$1996 (5,988,000 pixels) which took around 5 hours in 6 images. In our dataset FZ36147, its resolution is 4008$\times$5344 and contains 35 images, the pixels needed to be evaluated with \cite{dellepiane2012flow} will be 21,418,152 which is at least 5.57 times more than in their dataset with maximum resolution, and 6 times the number of images. Only with extremely powerful processing capable of handling such computations can their approach be applied, otherwise their method is not a feasible solution with our data set.

In general the state of the art methods \cite{eisemann2008floating}, \cite{dellepiane2012flow} are based on dense optical flow which is the main reason there is no possible comparison with our datasets.  

Even though our implementation has proven to be robust and reliable, some limitations still remain. The main one relates to the programming language for the acceleration of the computational time (from CPU to GPU programming). Also, in the evaluations performed, the maximum local displacement found was not large (10 pixels); but for other cases (e.g. images with higher resolution), the displacement can be bigger and in consequence the parameters for the template matching algorithm in Stage 3, have to be adjusted.

There are also some limitations related to lighting artifacts. Factors like highlights/shadows may complicate the estimation of the local error displacement, and inclusive mislead the motion to totally wrong values. Nevertheless , these drawbacks are shared with every method based on optical flow calculations.


\section{Conclusion}
We have proposed a semi-automatic 2D-3D registration pipeline capable to provide extremely accurate realistic results from a set of 2D uncalibrated images and a 3D object acquired through laser scanning. 

The main advantage of our pipeline is the generality, since no assumption is made about the geometric characteristics or shape of the object. Our pipeline is capable of handling registration with any kind of object, since the algorithm used is a brute force (SICP) which evaluates every single point and finds the best position. The only requirements needed are a set of 2D images containing sufficient overlapping information to be able to use the Structure from Motion (SfM) technique in stage 1; and a user intervention during stage 2 to locate the dense point cloud, coming from the scanner, closer to the one obtained by SfM, in order to provide the input that the Scale Iterative Closest Point (SICP) algorithm needs to converge. This user intervention during the second stage in our pipeline is what makes our approach semi-automatic. 

In conclusion, our main contribution is the local error correction algorithm in stage 3 which proved to be:
\begin{enumerate}
\item Robust: it works with low and high resolution images, as it considers the interest points (projected 3D points into the image plane) for the matching. Not even the state of the art \cite{dellepiane2012flow}, \cite{eisemann2008floating} is capable of dealing with as high resolution images as our algorithm.
\item Accurate: it finds the best possible matching for the small error displacements considering luminance and chrominance channels. Through the best match, it removes the unpleasant blurring artifacts and produces sharp results.  
\item Photo-realistic: with the point cloud generated by SFM \cite{snavely2006photo},\cite{snavely2008modeling} and the registration algorithm SICP \cite{zhu2010robust}, the color information from the 2D images is projected onto the 3D object transferring the appearance. 
\end{enumerate}

An interesting direction for future research would be to define a criterion with a respective threshold to identify the possible borders where the sharp results start to blur (in the cases where only some parts of the 3D object are visible with ghosting effects). This identification has to be based on the depth difference between the 2 registered point clouds, and probably a segmentation according to depth may help to optimized our proposed local error estimation algorithm.


\section*{Acknowledgments} 
Special thanks to Centre de Recherche et de Restauration des Musees de France (C2RMF) for the data provided to evaluate the pipeline. Thanks to J.H. Zhu, N.N. Zheng, Z.J. Yuan, S.Y. Du and L. Ma for sharing their SICP code implementation in Matlab.

\end{multicols}


\begin{thebibliography}{4}

\bibitem{el2002detailed} El-Hakim, S., Beraldin, J. A.: Detailed 3D reconstruction of monuments using multiple techniques. In Proceedings of the International Workshop on Scanning for Cultural Heritage Recording-Complementing or Replacing Photogrammetry, 58-64 (2002).

\bibitem{pollefeys2000automated} Pollefeys, M., Koch, R., Vergauwen, M., Van Gool, L.: Automated reconstruction of 3D scenes from sequences of images. ISPRS Journal of Photogrammetry and Remote Sensing, 55(4), 251-267 (2000).

\bibitem{fitzgibbon1998automatic} Fitzgibbon, A. W., Zisserman, A.: Automatic camera recovery for closed or open image sequences. In Computer Vision—ECCV'98, pp. 311--326. Springer Berlin Heidelberg (1998).

\bibitem{bernardini2001high} Bernardini, F., Martin, I. M., Rushmeier, H.: High-quality texture reconstruction from multiple scans. Visualization and Computer Graphics, IEEE Transactions on, 7(4), 318-332 (2001).

\bibitem{liu2006multiview} Liu, L., Stamos, I., Yu, G., Wolberg, G., Zokai, S.: Multiview geometry for texture mapping 2d images onto 3d range data. In Computer Vision and Pattern Recognition, 2006 IEEE Computer Society Conference, vol. 2, pp. 2293--2300. IEEE (2006).

\bibitem{dellepiane2008mapping} Dellepiane, M., Callieri, M., Ponchio, F., Scopigno, R.: Mapping highly detailed colour information on extremely dense 3D models: the case of david's restoration. In Computer Graphics Forum Vol. 27, No. 8, pp. 2178--2187. Blackwell Publishing Ltd (2008).

\bibitem{liu2005automatic} Liu, L., Stamos, I.: Automatic 3D to 2D registration for the photorealistic rendering of urban scenes. In Computer Vision and Pattern Recognition, 2005. CVPR 2005. IEEE Computer Society Conference on Vol. 2, pp. 137--143. IEEE (2005).

\bibitem{dellepiane2013} Dellepiane, M., Scopigno, R.: Global refinement of image-to-geometry registration for color projection on 3D models. In Digital Heritage International Congress, pp. 39--46. The Eurographics Association (2013).

\bibitem{li2009automatic} Li, Y., Low, K. L.: Automatic registration of color images to 3d geometry. In Proceedings of the 2009 Computer Graphics International Conference, pp. 21--28. ACM (2009).

\bibitem{dellepiane2012flow} Dellepiane, M., Marroquim, R., Callieri, M., Cignoni, P., Scopigno, R.: Flow-based local optimization for image-to-geometry projection. Visualization and Computer Graphics, IEEE Transactions on, 18(3), 463-474 (2012).
	
\bibitem{eisemann2008floating} Eisemann, M., De Decker, B., Magnor, M., Bekaert, P., De Aguiar, E., Ahmed, N., Sellent, A.: Floating textures. In Computer Graphics Forum, vol. 27, no. 2, pp. 409--418. Blackwell Publishing Ltd (2008).

\bibitem{takai2010harmonised} Takai, T., Hilton, A., Matsuyama, T.: Harmonised texture mapping. In Proc. of 3DPVT (2010).

\bibitem{liu19983d} Liu, A., Bullitt, E., Pizer, S. M.: 3D/2D registration using tubular anatomical structures as a basis. In Proc. Medical Image Computing and Computer-Assisted Intervention, pp. 952--963 (1998).

\bibitem{musialski2013survey} Musialski, P., Wonka, P., Aliaga, D. G., Wimmer, M., Gool, L., Purgathofer, W.: A survey of urban reconstruction. In Computer Graphics Forum, vol. 32, No. 6, pp. 146--177 (2013).

\bibitem{ip1996constructing} Ip, H. H., Yin, L.: Constructing a 3D individualized head model from two orthogonal views. The visual computer, 12(5), 254-266 (1996).

\bibitem{lensch2000automated} Lensch, H., Heidrich, W., Seidel, H. P.: Automated texture registration and stitching for real world models. In Computer Graphics and Applications, 2000. Proceedings. The Eighth Pacific Conference, pp. 317--452. IEEE (2000).

\bibitem{lensch2001silhouette} Lensch, H., Heidrich, W., Seidel, H. P.: A silhouette-based algorithm for texture registration and stitching. Graphical Models, 63(4), 245-262 (2001).

\bibitem{ikeuchi2007great} Ikeuchi, K., Oishi, T., Takamatsu, J., Sagawa, R., Nakazawa, A., Kurazume, R., Okamoto, Y.: The great buddha project: Digitally archiving, restoring, and analyzing cultural heritage objects. International Journal of Computer Vision, 75(1), 189-208 (2007).

\bibitem{kurazume2002simultaneous} Kurazume, R., Nishino, K., Zhang, Z., Ikeuchi, K.: Simultaneous 2D images and 3D geometric model registration for texture mapping utilizing reflectance attribute. Proc. Fifth ACCV, 99-106 (2002).

\bibitem{corsini2013fully} Corsini, M., Dellepiane, M., Ganovelli, F., Gherardi, R., Fusiello, A., Scopigno, R.: Fully automatic registration of image sets on approximate geometry. International journal of computer vision, 102(1-3), 91-111 (2013).

\bibitem{zhu2010robust} Zhu, J. H., Zheng, N. N., Yuan, Z. J., Du, S. Y., Ma, L.: Robust scaling iterative closest point algorithm with bidirectional distance measurement. Electronics letters, 46(24), 1604-1605 (2010).

\bibitem{callieri2002reconstructing} Callieri, M., Cignoni, P., Scopigno, R.: Reconstructing Textured Meshes from Multiple Range RGB Maps. In VMV, pp. 419--426 (2002).

\bibitem{callieri2008masked} Callieri, M., Cignoni, P., Corsini, M., Scopigno, R.: Masked photo blending: Mapping dense photographic data set on high-resolution sampled 3D models. Computers \& Graphics, 32(4), 464-473 (2008).

\bibitem{dellepiane2010improved} Dellepiane, M., Callieri, M., Corsini, M., Cignoni, P., Scopigno, R.: Improved color acquisition and mapping on 3d models via flash-based photography. Journal on Computing and Cultural Heritage (JOCCH), 2(4), 9 (2010).

\bibitem{pal1993review} Pal, Nikhil R., Sankar K. Pal.: A review on image segmentation techniques. Pattern recognition 26.9 (1993): 1277-1294.

\bibitem{naik2003hue} Naik, Sarif Kumar, C. A. Murthy.: Hue-preserving color image enhancement without gamut problem. Image Processing, IEEE Transactions on 12.12 (2003): 1591-1598.

\bibitem{gallo20143d} Gallo, A., Muzzupappa, M., Bruno, F.: 3D reconstruction of small sized objects from a sequence of multi-focused images. Journal of Cultural Heritage, 15(2), 173-182 (2014).

\bibitem{lowe2004distinctive} Lowe, D. G.: Distinctive image features from scale-invariant keypoints. International journal of computer vision, 60(2), 91-110 (2004).

\bibitem{arya1998optimal} Arya, S., Mount, D. M., Netanyahu, N. S., Silverman, R., Wu, A. Y.: An optimal algorithm for approximate nearest neighbor searching fixed dimensions. Journal of the ACM (JACM), 45(6), 891-923 (1998).

\bibitem{fischler1981random} Fischler, M. A., Bolles, R. C.: Random sample consensus: a paradigm for model fitting with applications to image analysis and automated cartography. Communications of the ACM, 24(6), 381-395 (1981).

\bibitem{snavely2006photo} Snavely, N., Seitz, S. M., Szeliski, R.: Photo tourism: exploring photo collections in 3D. ACM transactions on graphics (TOG), 25(3), 835-846 (2006).	

\bibitem{snavely2008modeling} Snavely, N., Seitz, S. M., Szeliski, R.: Modeling the world from internet photo collections. International Journal of Computer Vision, 80(2), 189-210 (2008).

\bibitem{furukawa2010towards} Furukawa, Y., Curless, B., Seitz, S. M., Szeliski, R.: Towards internet-scale multi-view stereo. In Computer Vision and Pattern Recognition (CVPR), 2010 IEEE Conference, pp. 1434--1441). IEEE (2010).

\bibitem{furukawa2010accurate} Furukawa, Y., Ponce, J.: Accurate, dense, and robust multiview stereopsis. Pattern Analysis and Machine Intelligence, IEEE Transactions on, 32(8), 1362-1376 (2010).

\bibitem{zhang1992iterative} Zhang, Z.: Iterative point matching for registration of free-form curves (1992).

\bibitem{aiger20084} Aiger, D., Mitra, N. J., Cohen-Or, D.: 4-points congruent sets for robust pairwise surface registration. In ACM Transactions on Graphics (TOG), Vol. 27, No. 3, p. 85. ACM (2008).

\bibitem{pighin2006synthesizing} Pighin, F., Hecker, J., Lischinski, D., Szeliski, R., Salesin, D. H.: Synthesizing realistic facial expressions from photographs. In ACM SIGGRAPH 2006 Courses (p. 19). ACM (2006).	


\end{thebibliography}
\end{document}